# Inferring Urban Mobility Interactions from Aggregated Dynamics

Yi Wang[1, 2, #], Jing Li[2, #, *], Jinliang Deng[3, *], Zhenghong Wang[4], Yizhi Zhang[1], Fan Zhang[1, *], Ivor W. Tsang[2], Yu Liu[1]

[1] Institute of Remote Sensing and Geographic Information System, School of Earth and Space Sciences, Peking University, Beijing, China
[2] Centre for Frontier AI Research, Institute of High-Performance Computing, Agency for Science, Technology and Research (A*STAR), Singapore
[3] Department of Computer Science and Engineering, Hong Kong University of Science and Technology, Hong Kong, China.
[4] College of Computer and Data Science, Fuzhou University, Fuzhou, China
[#] These authors have equal contributions
[*] Corresponding author(s): Jing Li (li_jing@a-star.edu.sg), Jinliang Deng (jinliangdeng9588@gmail.com), Fan Zhang (fanzhanggis@pku.edu.cn)

## Abstract

Real-time urban governance depends not only on knowing where people are, but on how they move between places, directional flows that could be conventionally resolved by tracking individuals through space—expensive to sustain and built on traces that are highly unique and readily re-identifiable. Here we show that this directional structure need not be observed to be known: aggregated counts which cities already collect retain enough information to reconstruct the temporal evolution of origin–destination (OD) matrix. Using an uncertainty-aware physics-informed framework (ODIN), we infer future OD flows from area-level counts alone across twelve mobility datasets from cities in the United States and China, reaching accuracy comparable to models that take historical OD matrices as input. Probabilistic modeling corrects the systematic underestimation of sparse, high-value corridors and yields calibrated predictions consistent with observed flows. Architectures that respect the generation-before-assignment logic of transport planning recover interactions more faithfully, indicating that location-level spatial heterogeneity should be preserved before pairwise interactions are reconstructed. Because inference requires only aggregated observations after training, recovering interactions this way reduces reliance on continuous individual-level tracking, pointing toward a more deployable and less exposure-heavy basis for real-time urban intelligence.

## Introduction

Spatiotemporal interactions are a fundamental organizing principle of complex systems, describing how entities exchange people, goods, information or risks across space and time[1]. They shape a wide range of dynamic processes, from global trade[2] and supply-chain connectivity[3] to epidemic transmission and evolving social relations[4]. Human mobility[5,6] represents one of the most direct and measurable forms of such interactions within cities. The continuous flux of human mobility underpins the resilience[4,7], productivity[8,9] and sustainable development of urban systems[10,11]. Real-time urban governance therefore increasingly depends on understanding not only where people are, but where they are moving from and to[12-14]. These directed, time-varying transitions are formalized as dynamic Origin-Destination (OD) flows, and interventions ranging from transport optimization[15] and crowd management[16] to infectious disease containment[17] all depend on knowing them. Yet the standard route to these flows runs through individual trajectories—costly to sustain, and intrusive by construction.

Learning-based paradigms have become the dominant route to dynamic OD inference, but they are typically formulated to predict OD flows from historical OD observations at the same granularity[18]. At deployment, they therefore presuppose a continuous supply of OD observations as input—a supply that the real world is doubly ill-equipped to provide. First, ubiquitous sensing infrastructures[19,20] (e.g., inductive loops and photoelectric sensors) typically provide only cross-sectional counting capabilities. This "count-but-not-track" mechanism captures local aggregate dynamics but fails to record individual end-to-end trajectories[21], leaving the underlying origin–destination interactions unobserved. Second, fine-grained OD observations from GPS or cellular signaling[22,23] are not the innocuous aggregates they appear to be. At the resolutions required for real-time inference, a large fraction of OD–time cells contain a single trip (Fig. 1a and Fig. 6), and such cells are individual records in aggregate clothing. Montjoye et al. have shown that traces of this kind are highly unique, with as few as four spatiotemporal points sufficing to identify an individual from a population of millions with 95% accuracy[24]. These limitations motivate a shift from directly observing OD flows to asking whether spatially aggregated dynamics[25,26], which are more widely accessible and already routinely collected, still retain recoverable fine-grained information about latent mobility interactions.

To examine this recoverability, we study coarse-to-fine dynamic OD prediction. This problem is severely under-constrained[27-29]: recovering high-dimensional OD interactions from low-dimensional aggregate counts resembles inferring a joint distribution from its marginals, so that at any single instant the same aggregate observations are consistent with vastly different underlying OD structures. This difficulty is compounded by the temporal dynamics, as periodicity, volatility and sparsity at fine resolutions jointly amplify the ambiguity. Yet aggregation does not erase interaction structure entirely: temporal and spatial regularities may leave indirect but recoverable signatures of the latent flows[30-32], an assumption our study is designed to test.

Classical physical principles, such as maximum entropy[33,34], can in principle constrain OD recovery toward physically plausible interaction structures, but their canonical realizations, such as gravity model[33] or radiation model[35] assume dense and stationary interactions. Real-time OD flows violate such assumptions as they shift strongly across time and remain highly sparse. Standard deterministic learning faces a complementary obstacle rooted in the long-tail nature of mobility[5,36], where abundant low-value flows dominate the sparse, high-value corridors that carry most operational significance. This imbalance pulls estimates toward the low-volume majority and systematically underestimates high-value flows so that interactions become hardest to recover precisely where accuracy matters most, during mass gatherings or epidemic spread (Fig. 1b).

To test whether this recoverable information is truly present, we develop ODIN, an uncertainty-aware physics-informed framework, as a probe. Across twelve mobility datasets from cities in the United States and China, we find that aggregated dynamics retain recoverable information about latent mobility interactions: from area-level counts alone, ODIN infers future OD flows with accuracy comparable to models that observe full historical OD. Modeling OD flows probabilistically further corrects the systematic underestimation of sparse high-value corridors and yields calibrated predictions consistent with observed flow distributions. Recovery also depends on modeling order: architectures aligned with the generation-before-assignment logic of transport planning recover interactions more faithfully, indicating that location-level spatial heterogeneity must be preserved before pairwise interactions are reconstructed. Because inference then requires only aggregated observations, the need for individual-level data shifts from continuous operation to a one-time training stage, substantially reducing both the annual acquisition burden and the exposure of sparse, potentially identifiable records.

## Results

### Overview Task formulation and ODIN framework

In this study, the urban area is defined as a graph $\mathcal{G}(\mathcal{V},\mathcal{E})$, where spatial units (e.g., traffic zones or census tracks) are represented as vertices $\mathcal{V} \in \mathbb{R}^{N}$, and the connectivity between these units is represented as edges $\mathcal{E} \in \mathbb{R}^{N\times N}$. Unlike traditional forecasting tasks that extrapolate future states using historical observation within the same granularity, our task is to infer fine-grained edge-level OD flow from coarse-grained node-level aggregated dynamics (i.e., inflow, outflow and overall flow).

Let $X_{in} = \{X_{t-q+1}^{in}, X_{t-q+2}^{in}, \dots, X_{t}^{in} | \mathcal{G}\} \in \mathbb{R}^{N\times T_{in}\times 1}$ and $X_{out} = \{X_{t-q+1}^{out}, X_{t-q+2}^{out}, \dots, X_{t}^{out} | \mathcal{G}\} \in \mathbb{R}^{N\times T_{in}\times 1}$ denote the historical observed inflow and outflow intensity signals on $N$ nodes over $T_{in}$ time steps. Here, $q$ denotes the number of time steps within the observation window $T_{in}$. The target is the future OD flow matrix, denoted as $Y_{OD} = \{Y_{t+1}, Y_{t+2}, \dots, Y_{t+p} | \mathcal{G}\} \in \mathbb{R}^{N\times N\times T_{out}\times C}$, representing the movement magnitude between any pair of nodes in every future time step $p$. Therefore, the proposed task, i.e., coarse-to-fine dynamic OD flow prediction, can be formally defined as follows: Given the historical coarse-grained flow observations $X_{obs} = \{X_{in}, X_{out} | \mathcal{G}\}$, the objective is to learn a mapping function $F(\cdot)$ to predict the future fine-grained OD matrices $Y_{OD}$:

$$\underbrace{\{X_{in}, X_{out} | \mathcal{G}\}}_{X_{obs}} \xrightarrow[\mathcal{F}]{} \underbrace{\{Y_{t+1}, Y_{t+2}, \dots, Y_{t+p} | \mathcal{G}\}}_{Y_{OD}}, \tag{1}$$

To recover latent OD interactions from these aggregated inputs, we develop ODIN (Fig. 2), an uncertainty-aware, physics-informed framework built on two ideas. A spatial and temporal transformer backbone first captures how each location's mobility state evolves in time and co-varies across space, yielding representations shared by two downstream modules. The Temporally Resolved Gravity Model (TRGM) supplies a physical prior for structural recovery: rather than assuming a single static interaction pattern, it lets gravity-type structure vary over time, anchoring plausible OD structure within an otherwise intractable solution space. The Disentangled Gaussian Mixture Model (DGMM) then addresses distributional imbalance: because the long-tailed nature of OD flows leads deterministic estimators to systematically underestimate sparse, high-value corridors, DGMM separates reducible (epistemic) from irreducible (aleatoric) uncertainty and uses the former to recalibrate these underestimated flows. ODIN is then trained by maximizing likelihood under regularization that stabilizes the mixture and its training dynamics (See Methodology). Together, these components make ODIN a probe designed to extract as much recoverable structure as the aggregated signal permits—the basis for the analyses that follow.

### Baselines and evaluation metrics

To establish how much of the OD structure existing methods can recover under this setting, we select benchmark methods spanning time-series forecasting, OD flow generation, spatiotemporal forecasting and OD flow forecasting, ranging from classical to state-of-the-art approaches, for comparison with ODIN. Specifically, these include: (1) time-series forecasting methods: DLinear[37] and iTransformer[38]; (2) OD flow generation methods: Deep Gravity[39], GEML[40], ODGN[41], and GSTE-DF[42]; (3) spatiotemporal forecasting methods (deterministic forecasting): STGCN[43], STID[44], and STAEformer[45]; (4) spatiotemporal forecasting methods (uncertainty-aware forecasting): STNB[46], STGS[46], STTD[47], and UQGNN[48]; (5) OD flow forecasting methods: GEML*[49], ODCRN[50], HIAM[51], CAMLO[52], ODMixer[53], and PDMTSOD[54].

Because coarse-to-fine dynamic OD flow prediction differs from other existing paradigms (refer to task settings comparison in Supplementary), existing methods must be repurposed for this setting. We evaluate all baselines under a unified information constraint: only historical node-level aggregated dynamics, including inflow and outflow, are used to predict future fine-grained OD flows. For methods originally relying on auxiliary or multi-source information, we retain their core spatiotemporal modeling components but instantiate them with the same node-level mobility signals and temporal embeddings as ODIN. This avoids unfair comparison caused by unequal information access while preserving the essential modeling capacity of each baseline. According to their primary modeling objects, we further group the repurposed baselines into two categories:

- **Node-first modeling methods.** These methods first encode node-level temporal signals and are therefore naturally compatible with our aggregated inputs. Their original encoder is kept unchanged. To predict OD flows, we convert the learned origin and destination node representations into edge representations through an interaction layer at the decoding stage, followed by a multilayer perceptron (MLP) projection to generate future OD flows.
- **Edge-first modeling methods.** These methods are originally designed to model historical OD matrices. Since such fine-grained inputs are unavailable in our setting, we construct their initial edge representations from paired origin–destination embeddings derived from node-level inflow and outflow signals during encoding stage. The original edge-oriented forecasting modules are then retained for subsequent modeling.

To evaluate prediction performance, we employ three regression metrics: root mean square error (RMSE), mean absolute error (MAE), and weighted mean absolute percentage error (WAPE). Notably, we adopt WAPE instead of MAPE to avoid bias caused by zero values in the denominator, given the prevalence of zeros in OD data. Furthermore, to assess the reliability of uncertainty quantification, we introduce two additional metrics for models capable of uncertainty estimation: Continuous Ranked Probability Score (CRPS) and Jensen-Shannon Divergence (JSD). The detailed mathematical formulations of these metrics are provided in the Supplementary section Experiment settings.

**How far aggregated inputs support OD recovery across methods**

*(1) Recovery from aggregated inputs matches OD-based models*

Across all twelve datasets, methods restricted to aggregated inputs recover fine-grained OD flows to a degree that varies widely, and ODIN recovers them most fully: it matches, and in several cases exceeds, models that observe complete historical OD matrices as input, improving over the strongest aggregated-input competitor by up to 15% in RMSE (Fig. 3a, Extended Data Fig. 1). That a model reading only node-level inflow and outflow can rival methods with direct OD access indicates that much of the recoverable structure is already latent in the aggregated signal—and that ODIN extracts it more completely than the other repurposed methods. Beyond aggregate error, the flows ODIN recovers preserve the observed temporal structure of mobility, tracking day-of-week variation with the smallest weekend fluctuation among all compared models (Fig. 3b, c).

Under the proposed coarse-to-fine prediction setting, time-series forecasting methods and OD flow generation methods are generally less competitive. Time-series models mainly capture temporal variations but lack explicit modeling of origin–destination interactions, whereas OD generation models focus on spatial allocation but are less effective in tracking short-term temporal dynamics. These complementary failures show that recoverability from aggregated inputs is unlocked only when temporal dependencies and spatial interactions are modeled jointly—neither alone suffices.

Node-level spatiotemporal forecasting methods (both deterministic and uncertainty-aware), particularly STAEformer and UQGNN, achieve stronger performance than most other baselines. GSTE-DF, the only dynamic OD generation baseline, also performs competitively. By contrast, conventional OD forecasting methods do not show a clear advantage in this setting. These methods are originally designed to exploit historical OD matrices, where observed edge-level sparsity provides direct cues for future prediction. In the proposed task, such fine-grained OD observations are unavailable. Therefore, directly prioritizing edge-level modeling from reconstructed edge features may introduce redundant or noisy interactions. These results suggest that learning effective node-level representations is crucial for decoding latent interactions from aggregated dynamics. We further examine this issue in the following section by comparing node-first and edge-first modeling under different input granularities.

The performance of ODIN suggests that aggregated dynamics can be decoded into fine-grained mobility interactions when node-level spatiotemporal representation learning is combined with physics-informed and uncertainty-aware OD dynamics inference. Specifically, TRGM constrains the reconstruction of edge-level interactions from node-level dynamics, while DGMM improves prediction reliability under imbalanced OD flow distributions.

*(2) ODIN holds up where recovery is hardest*

Recovery is not uniformly difficult. Prediction error rises with OD-flow magnitude across all methods (Fig. 4a), confirming that sparse, high-value corridors—the interactions that matter most operationally—are the hardest to infer from aggregated observations. This difficulty is intrinsic to the task: aggregate counts constrain such rare flows only weakly. What distinguishes ODIN is that its error grows most slowly in exactly this regime, recovering high-value interactions that abundant low-value flows tend to mask. The same holds across time and space: ODIN forms the lower error bound across most hours of the day (Fig. 4b) and produces fewer high-error links and more spatially coherent predictions than competing methods (Fig. 4c).

In the temporal view, we compare the average hourly RMSE of different models. ODIN consistently forms the lower bound across most time periods, showing that its advantage is not restricted to specific hours but remains stable under varying daily mobility patterns. In the spatial view, flow maps of WAPE further reveal that ODIN produces fewer high-error OD links and more spatially coherent predictions. These results indicate that ODIN reduces prediction errors not only on average, but also across high-value flows, temporal variations and spatial interaction patterns.

*(3) Uncertainty improves the recoverable reliability*

To evaluate whether probabilistic modeling improves the reliability of interaction recovery, we compare all uncertainty-aware methods using CRPS and JSD. A reliable uncertainty estimate should be both accurate and well calibrated: lower CRPS indicates that the predicted probabilistic distribution better matches the observed OD flow, while lower JSD reflects smaller divergence between the predicted and empirical distributions. As shown in Extended Data Table 1, ODIN consistently outperforms other uncertainty-aware baselines across these metrics, demonstrating that it provides more reliable probabilistic recovery of OD interactions in addition to lower prediction errors.

We further visualize the epistemic and aleatoric uncertainty of ODIN across temporal and spatial dimensions (Fig. 4d–e). Aleatoric uncertainty remains relatively stable, reflecting irreducible randomness in mobility demand. In contrast, epistemic uncertainty varies more strongly with flow magnitude and prediction error, and spatially concentrates on OD links where prediction is more difficult. This pattern suggests that epistemic uncertainty captures recoverable model uncertainty caused by imbalanced observations. Therefore, epistemic uncertainty provides a useful calibration signal for recovering high-value OD interactions, where deterministic models tend to underestimate flows.

**Aggregated dynamics retain sufficient information for interaction inference**

The central claim of this study is that aggregated urban dynamics retain recoverable information about mobility interactions. A key test of this claim is whether reducing input granularity from historical OD matrices to node-level aggregate counts substantially weakens OD recovery. To examine this, we further conduct controlled experiments under three input settings: historical OD matrices, node-level inflow/outflow, and node-level total flow.

Recovery accuracy degrades only marginally along this sequence. Even when the input is stripped to a single total-flow count per location, discarding the directional split between inflow and outflow entirely, ODIN recovers OD structure at accuracy close to that obtained from full OD observations (Fig. 5a). The signal required to reconstruct directional interactions therefore does not depend on observing those interactions, or even on their directional aggregates; it persists in the coarsest summary of local activity. A non-learning reference that allocates the same aggregated counts by historical averages and a static gravity model (HA + Gravity, Fig. 5a) recovers OD far less accurately, confirming that this recoverability comes from modeling the dynamics of the aggregate signal rather than from the aggregate counts alone. This robustness is not shared by all architectures: node-first methods remain stable as the input coarsens, whereas edge-first methods built to operate on OD matrices degrade sharply once true OD input is withheld, with the strongest OD-forecasting baseline, ODMixer, losing much of its accuracy in this regime. The underlying reason for this node-first advantage is examined in the next section.

Together these results settle the central question of this study. The fine-grained interactions that cities depend on are not lost to aggregation; they remain latent in the coarse dynamics already being collected, and can be recovered without ever observing OD directly. Aggregated observations are therefore not merely low-resolution substitutes for trajectory data. They carry enough structure to reconstruct directional interactions while removing the need for individual-level tracking at deployment, an implication we quantify later.

**Generation-before-assignment logic explains the advantage of node-first modeling**

The node-first advantage seen above raises a sharper question: is it merely a matter of input information, or does the modeling order itself shape what can be recovered? To separate the two, we compare the original node-first ODIN against an edge-first variant that shares all components but reverses the stage at which OD inference is performed.

For a fair comparison, the edge-first variant retains the main components of ODIN but changes the stage at which OD inference is performed. In the original ODIN, node-level representations are first learned and then converted into edge-level OD representations during decoding. In the edge-first variant, we construct edge-level representations immediately after the embedding layer, before the Transformer blocks, so that subsequent modules operate directly on origin–destination pairs. The remaining components are kept unchanged.

The results show that the edge-first variant underperforms the original node-first ODIN in most input settings (Fig. 5b). Although providing historical OD matrices improves the edge-first variant, the gain remains limited. This indicates that the performance gap is not only caused by input information, but is also related to the learning preference induced by the modeling paradigm. In this task, early edge construction may introduce redundant pairwise interactions, whereas node-first modeling preserves compact and informative representations before OD inference.

To further understand this difference, we extract latent features from the penultimate layer of both variants and visualize them using t-SNE[55] (Fig. 5c). The node-first model produces a smoother and more structured latent space[56], whereas the edge-first model shows a more fragmented distribution. In deep regression tasks, smoother latent representations are often associated with better generalization[57], because nearby samples tend to produce more consistent predictions. This observation is consistent with the stronger predictive performance of node-first modeling.

The two modeling paradigms also exhibit different preferences in temporal and spatial dimensions. Edge-first modeling tends to separate features by forecast horizons, suggesting a stronger emphasis on temporal heterogeneity. By contrast, node-first modeling produces clearer spatial clustering, i.e., nodes with similar flow levels are grouped more coherently in the latent space. This suggests that, for coarse-to-fine OD prediction, preserving spatial heterogeneity at the node level is more beneficial than directly modeling all pairwise edge interactions from the beginning.

This finding is consistent with the sequential logic of classical transportation planning framework[58]. In the conventional four-step framework, trip planning at locations precedes the distribution and assignment of human mobility flows[59]. In other words, spatial units and their functional attributes provide the basis for mobility generation[39], while OD flows are the resulting interactions[59]. From this perspective, node-first modeling better aligns with the generative process of human mobility. Specifically, it first learns the location-level causes of mobility and then reconstructs the edge-level effects. Edge-first modeling, although effective when historical OD matrices are available, may be less suited to setting of aggregated-input recovery because it starts from pairwise effects before sufficiently learning their node-level origins. More broadly, this suggests a principle for recovering interactions from aggregated signals: represent the state of places before inferring the flows between them. Where the generative order of mobility is respected, recovery is more faithful.

**Reducing data acquisition burden and privacy exposure**

A practical implication of recovering OD interactions from aggregated dynamics is the reduced dependence on trajectory-level data acquisition and exposure. To quantify this advantage, we conducted an annualized cost analysis using literature-calibrated unit prices. Area-level aggregated sensing was parameterized using volume-only traffic count prices of US$375–725 per station for 7 days[60], equivalent to US$19,875–38,425 per station-year over the 366 days of 2024. Individual-level trajectory acquisition was parameterized using a survey-cost range from US$25 per GPS-instrumented participant[61] for a 3-day tracking window to US$350 per completed survey unit[62]. Assuming 50 trajectory records per participant per day, this corresponds to US$0.167–2.333 per trajectory record. We then combined these unit prices with the number of required counting stations or participants in each dataset to estimate annual acquisition costs (Fig. 6). Across all datasets, area-level aggregated sensing costs approximately US$4.85–9.38 million, whereas trajectory-based acquisition costs approximately US$18.38–257.37

million, corresponding to an estimated 73.6–96.4% reduction in annual acquisition burden. The lower bound of trajectory acquisition uses US$25 per participant from a volunteer GPS-survey incentive. This is a highly conservative estimate. Real-world purchase or continuous collection of individual trajectory data would likely cost more, so the actual cost advantage of area-level sensing may be larger.

We further quantified privacy exposure using cell-level uniqueness and *k*-anonymity risk[63]. OD-level observations were evaluated over origin–destination–time cells, whereas aggregated observations were evaluated over node–time cells. Across datasets, OD flows show an average unique-cell rate ($k$=1) of 43.9% and low-3 exposure ($k$=3) of 61.8%. In contrast, inflow/outflow aggregation reduces these values to 16.3% and 24.0%, corresponding to reductions of 72.0% and 67.9%, respectively. Total-flow aggregation further reduces the average unique-cell rate and low-3 exposure to 12.1% and 18.2%. These results indicate that area-level aggregated observations can reduce both annual acquisition burden and the exposure of sparse, potentially distinguishable mobility records, although they do not provide a formal privacy guarantee. A distinction is essential here. The privacy exposure we quantify concerns the acquisition of OD data, where each fine-grained cell traces back to real individuals whose trajectories were recorded. ODIN's recovered OD flows are model estimates inferred from aggregate counts; no individual trajectory is collected or held at inference, and a recovered single-trip cell corresponds to no identifiable person. The reduction in exposure therefore arises at the sensing stage, not merely in the choice of what to output.

## Discussion

Fine-grained mobility interactions are among the most useful yet most costly and privacy-sensitive urban data to collect. Here we have shown that this directional structure need not be observed to be known: aggregated inflow and outflow, which cities already collect and which reveal nothing about who moves where, retain enough information to recover future OD flows. Once trained, ODIN draws only on these aggregated counts, substantially reducing the recurring cost of data acquisition and the exposure of sparse, potentially identifiable records. Cities can thus infer fine-grained directional interactions from accessible, already-collected aggregate sensing, rather than from the continuous collection of individual trajectories or full OD matrices.

This capability is important because many urban interventions require directional mobility information, but not necessarily individual-level tracking. During epidemic response, inferred OD dynamics can help identify likely transmission corridors and support the placement of targeted screening or inspection stations[4]. In traffic management, direction-specific OD dynamics inference can inform adaptive signal control for specific road segments, turning movements or commuting corridors[64]. For mass events, emergency evacuation and public transport operations, the same framework could help anticipate where mobility demand will emerge and where flows are likely to concentrate. In each case, ODIN supports finer-grained management without deploying new infrastructure to continuously track individuals.

Methodologically, our contribution is a reframing: we treat dynamic OD prediction not as forecasting from same-granularity history, but as recovering latent interactions under an information constraint, where the observable signal (aggregate counts) is coarser than the target (directional flows). This shift turns a data-hungry forecasting problem into one of extracting the interaction structure already latent in aggregate dynamics, using physical priors to constrain plausible OD allocations and uncertainty modeling to recover the sparse, high-value flows that dominate operational risk. Framing the problem this way is what allows recovery to proceed without OD inputs at all.

The experiments also reveal a modeling principle for coarse-to-fine interaction recovery that is relevant beyond this specific architecture. Models that better follow the generation-before-assignment logic[58] in transportation planning achieve stronger reconstruction performance. This suggests that, when historical OD observations are unavailable, it is more effective to first preserve the dynamic states and spatial heterogeneity of locations, and then reconstruct their pairwise interactions. This points to the central role of place representation: the model must first learn how locations differ in their mobility potential before inferring how they interact over time. In practical terms, node-level dynamics describe the mobility potential of places, whereas OD flows represent realized interactions between them. Respecting this generative order aligns model design with classical transportation reasoning and offers a principle for recovering interactions from any aggregated signal: represent the parts before inferring the whole.

More broadly, this work points to a responsible direction for geospatial artificial intelligence. Fine-grained interaction knowledge underpins many urban functions, yet acquiring it need not depend on intrusive or expensive individual-level tracking. A distinction is central to this claim. The exposure we reduce is that of data acquisition, where each fine-grained OD cell traces back to a real person whose movement was recorded. ODIN's recovered flows, by contrast, are model estimates inferred from aggregate counts, and a recovered single-trip cell corresponds to no identifiable individual. Recovery therefore shifts reliance away from collecting and retaining real trajectories, rather than reproducing that risk in its outputs. This is not a formal anonymity or differential-privacy guarantee, but an operational reduction in the need to continuously acquire and hold individual-level records. Aggregated observations can, under some conditions, still be susceptible to reconstruction[28,29].

Several limitations remain. First, our approach recovers OD without OD inputs only at deployment; training still requires ground-truth OD as supervision. The method therefore suits settings where historical OD is available once to train a model that is then run on aggregate counts alone, rather than settings with no OD access at all. Second, recoverability is bounded by the information the aggregate signal carries: in sparse, highly volatile regimes, aggregate counts constrain directional flows only weakly, and recovery accuracy is correspondingly limited—a property of the task rather than of any particular model. Third, although our datasets span the United States and China, broader cross-regional generalization remains to be tested, as mobility systems vary with urban form, infrastructure, land use and policy. Finally, the framework deliberately relies only on aggregate dynamics and distance; incorporating non-sensitive context (road networks, land use, weather, events, transit schedules) may improve robustness under atypical conditions, and multimodal urban foundation models may help probe the upper bound of what is recoverable, though reliable OD-supervised data for such pretraining remain scarce.

## Methodology

Our proposed method consists of four main components: node-level dynamics embedding, spatiotemporal representation learning, temporally resolved gravity-based edge reconstruction and disentangled gaussian mixture model.

### Node-level dynamics embedding

The first step of ODIN is to embed raw node-level mobility observations into a latent representation for subsequent spatiotemporal modeling. At each historical time step, each spatial unit contains two aggregated signals, inflow and outflow, denoted as $X_{obs} \in \mathbb{R}^{N \times T_{in} \times 2}$. These signals describe the local mobility state of each unit, with inflow reflecting attraction intensity and outflow reflecting production intensity. We project them into a high-dimensional mobility feature embedding through a linear layer:

$$E_f = W_f X_{obs} + b_f, \tag{2}$$

Human mobility also exhibits strong temporal regularities. We therefore introduce learnable temporal embeddings to encode time-of-day and day-of-week information, rather than treating timestamps as fixed periodic patterns. Specifically, daily period embeddings $E_d \in \mathbb{R}^{D\times C_t}$ and weekly period embeddings $E_w \in \mathbb{R}^{7\times C_t}$ provide temporal context for learning daily and weekly mobility rhythms. In addition, adaptive embeddings $E_a \in \mathbb{R}^{N\times T\times C_a}$ are introduced as learnable parameters to capture latent spatial heterogeneity and recurring spatiotemporal patterns not explicitly represented by inflow and outflow.

The final initial representation is obtained by concatenating these components and projecting them through a linear transformation:

$$Z^0 = W_o(E_d \oplus E_w \oplus E_a \oplus E_f) + b_o, \tag{3}$$

where $\oplus$ denotes concatenation, $W_f \in \mathbb{R}^{2\times C_f}$, $W_o \in \mathbb{R}^{(2C_t+C_a+C_f)\times C_{hid}}$, $b_f \in \mathbb{R}^{C_f}$ and $b_o \in \mathbb{R}^{C_{hid}}$ are learnable parameters.

**Spatiotemporal dependencies modeling**

After obtaining the initial node representation $Z^0$, we use Transformer blocks to learn dynamic dependencies along the temporal and spatial dimensions. This module differs from the embedding layer: the embedding layer only maps observed signals and temporal identifiers into a latent space, whereas the transformer updates these representations by modeling interactions across time steps and spatial units.

Each block follows the vanilla transformer architecture, consisting of multi-head self-attention $\mathcal{F}_{MHSA}$, feed-forward networks $\mathcal{F}_{FFN}$ and layer normalization $\mathcal{F}_{LN}$. In our setting, temporal attention captures dependencies among historical mobility states, such as short-term fluctuations and periodic changes, while spatial attention captures coordinated variations among urban units. For the $l$-th layer ($l \in [1, L]$), the computation is formulated as:

$$Z_{hid}^{(l)} = \mathcal{F}_{LN}(Z^{(l-1)} + \mathcal{F}_{MHSA}(Z^{(l-1)})), \tag{4}$$

$$Z^{(l)} = \mathcal{F}_{LN}(Z_{hid}^{(l)} + \mathcal{F}_{FFN}(Z_{hid}^{(l)})), \tag{5}$$

where $Z^{(l)}$ denotes the output of the $l$-th transformer block. Following this structure, we first apply the transformer along the temporal dimension and then along the spatial dimension to obtain the final node-level spatiotemporal representation $Z_{ST}^{(L)}$, which is passed to the subsequent edge reconstruction module.

**Spatiotemporal interaction reconstruction**

The spatiotemporal transformer captures node-level dynamics, while OD prediction requires edge-level representations between origin–destination pairs. This node-to-edge mapping is intrinsically underdetermined, because many possible OD matrices can correspond to the same aggregated inflow and outflow observations. To constrain this solution space, we introduce a gravity-informed inductive bias[65] for spatial interaction reconstruction.

The classical gravity model assumes that interaction intensity increases with origin and destination importance and decreases with geographical distance[12]. In mobility systems, this formulation provides a physically interpretable prior for estimating potential interactions between locations based on maximum entropy theory[34]. Its standard form and logarithmic transformation[66] are:

$$T_{ij} = G\frac{{P_i}^{\alpha_1}{P_j}^{\alpha_2}}{D^{\beta}}, \tag{6}$$

$$\log\left(T_{ij}\right) = logG + \alpha_1 logP_i + \alpha_2 logP_j - \beta logD, \tag{7}$$

where $T_{ij}$ denotes the interaction intensity from location $i$ to location $j$, $P_i$ and $P_j$ denote the importance of the origin and destination, $D$ is their geographical distance, and $G, \alpha_1, \alpha_2, \beta$ are learnable scaling parameters.

However, classic gravity models are mainly designed for static and relatively dense interaction patterns. In our setting, OD flows are dynamic, sparse and highly imbalanced. We therefore propose the Temporally Resolved Gravity Model (TRGM) to stabilize edge representation learning by combining point-to-point local interaction and patch-to-patch temporal trend interaction. The point-to-point branch estimates local origin–destination attraction from node representations, while the patch-to-patch branch compares historical inflow and outflow patterns to capture broader temporal compatibility between spatial units. These two interaction features are further integrated with a distance-decay kernel to construct edge-level representations for future OD prediction.

Specifically, the transformer output $Z_{ST}^{(L)}$ is first mapped to the prediction horizon to obtain $Z$. The point-to-point interaction $I_{pt}$ is then computed from two projected node representations, corresponding to origin-side production and destination-side attraction. For patch-to-patch trend modeling, we apply the Fast Fourier Transform to inflow and outflow sequences and use Mahalanobis distance to compare their dominant temporal components[67,68]. $I_{tr1}$ measures the compatibility between upstream outflow and downstream inflow patterns[69], while $I_{tr2}$ measures the reverse pattern similarity. The two trend features are fused into $I_{tr}$. Finally, $I_{pt}$, $I_{tr}$, and the gaussian distance-decay kernel[43] $w$ are combined to obtain the final edge-level interaction representation $I$:

$$Z = W_{e2}(W_{e1}Z_{ST}^{(L)} + b_{e1}) + b_{e2} \tag{8}$$

$$I_{pt} = (W_a Z + b_a)^{\top}(W_r Z + b_r) \tag{9}$$

$$I_{tr1} = \left(\mathcal{F}_{fft}(X_{in}) - \mathcal{F}_{fft}(X_{out})\right)^{\top} Q_1 \left(\mathcal{F}_{fft}(X_{in}) - \mathcal{F}_{fft}(X_{out})\right) \tag{10}$$

$$I_{tr2} = \left(\mathcal{F}_{fft}(X_{out}) - \mathcal{F}_{fft}(X_{in})\right)^{\top} Q_2 \left(\mathcal{F}_{fft}(X_{out}) - \mathcal{F}_{fft}(X_{in})\right) \tag{11}$$

$$I_{tr} = W_{t2}(W_{t1}(I_{tr1} \oplus I_{tr2}^{\top}) + b_{t1}) + b_{t2} \tag{12}$$

$$w = exp(-\frac{D^2}{\tau^2}) \tag{13}$$

$$I = W_{g2}(W_{g1}\left(I_{pt} \oplus I_{tr} \oplus w\right) + b_{g1}) + b_{g2}, \tag{14}$$

where $\mathcal{F}_{fft}(\cdot)$ denotes the Fast Fourier Transform, which extracts dominant temporal components for more stable sequence comparison. $Q_1, Q_2$ is the learnable matrix used in the Mahalanobis distance. $w$denotes the Gaussian distance-decay kernel derived from the geographical distance matrix $D$. And $W_*, b_* (* \in e1, e2, t1, t2, g1, g2)$ denote learnable projection parameters with compatible dimensions. The resulting $I$ serves as the edge-level representation for the subsequent probabilistic OD decoder.

**Uncertainty-aware spatiotemporal interaction decoding**

Based on the edge-level interaction representation $I$, we further decode future OD flows in a probabilistic manner. Dynamic OD flows are sparse and long-tailed, and a deterministic decoder tends to underestimate rare but important high-value flows. We therefore introduce the Disentangled Gaussian Mixture Model (DGMM), which first models the predictive distribution of OD flows and then separates epistemic uncertainty from aleatoric uncertainty for reliability-aware calibration.

Specifically, we first use a Gaussian Mixture Model (GMM) to avoid assuming a single parametric form for the target distribution. Given the edge representation $I$, a multilayer decoder outputs the mixture parameters $\{\mathcal{N}(\mu,\sigma^2),\pi\}_{k=1}^{K}$:

$$H = W_{i1}I + b_{i1}, \tag{15}$$

$$\pi = \mathcal{S}(W_{pi}H + b_{pi}), \tag{16}$$

$$\mu = \theta(W_{mu}H + b_{mu}), \tag{17}$$

$$\sigma^2 = exp(W_{sig}H + b_{sig}), \tag{18}$$

where $H$ denotes the hidden decoding feature. $\pi$, $\mu$and $\sigma^2$ are the mixture coefficient, mean and variance of the Gaussian components, respectively. $\mathcal{S}(\cdot)$ denotes the Softmax function, $\theta(\cdot)$denotes the Softplus function for non-negative OD estimates, and $exp(\cdot)$ ensures positive variance. $W_*$ and $b_*$ ($* \in \pi,\mu,\sigma$) are learnable decoder parameters with compatible dimensions.

Although the GMM decoder provides a full predictive distribution, directly using its mean can still lead to biased estimates for sparse high-value OD flows. This is because probabilistic learning under imbalanced OD distributions may assign weaker effective gradients to samples with large predicted variance. For a Gaussian likelihood, the gradient with respect to the mean is proportional to $(\mu - y)/\sigma^2$, indicating that samples with larger predicted variance contribute smaller effective gradients. When this variance mainly reflects epistemic uncertainty caused by sparse or imbalanced observations, the model may underfit rare high-value OD flows. Epistemic uncertainty therefore provides a useful signal for identifying underrepresented regions of the target distribution and increasing their calibration strength.

However, the predictive uncertainty of a standard GMM mixes two different sources: aleatoric uncertainty, which reflects irreducible randomness in mobility demand, and epistemic uncertainty, which reflects model uncertainty caused by insufficient or imbalanced observations. Since only epistemic uncertainty is reducible and informative for correcting model bias, we separate it from total predictive uncertainty using Jensen–Rényi Divergence (JRD). Specifically, JRD measures the gap between the uncertainty of the whole mixture and the average uncertainty of individual Gaussian components, and this gap is interpreted as epistemic uncertainty:

$$\varphi_{epi} = \underbrace{\mathcal{F}_{JRD}(\{\mathcal{N}(\mu,\sigma^2)\}_{k=1}^{K})}_{Epistemic\ uncertainty} = \underbrace{H_\alpha\left(\sum_{k=1}^{K}\frac{1}{K}\mathcal{N}_k(\mu,\sigma^2)\right)}_{Total\ uncertainty} - \underbrace{\sum_{k=1}^{K}\frac{1}{K}H_\alpha\big(\mathcal{N}_k(\mu,\sigma^2)\big)}_{Aleatoric\ uncertainty}, \tag{19}$$

Directly computing the entropy of a Gaussian mixture is generally intractable, and Monte Carlo estimation is computationally expensive for real-time OD prediction. We therefore adopt a closed-form approximation of JRD based on quadratic Rényi entropy ($\alpha = 2$), which avoids repeated sampling:

$$H_\alpha(Z) = \frac{1}{1-\alpha} log \int p(z)^\alpha dz, \tag{20}$$

$$\varphi_{epi} \approx -\log\left[\frac{1}{K^2}\sum_{i=1}^{K}\sum_{j=1}^{K}\mathfrak{D}_{ij}\right] + \frac{1}{K}\sum_{i=1}^{K}\log[\mathfrak{D}_{ii}], \tag{21}$$

$$where \quad \mathfrak{D}_{ij} = \frac{1}{\sqrt{\sigma_i^2 + \sigma_j^2}} exp(-\frac{(\mu_i - \mu_j)^2}{2(\sigma_i^2 + \sigma_j^2)}), \tag{22}$$

where $H_\alpha(\cdot)$ denotes Rényi entropy, $\mathfrak{D}_{ij}$ measures the overlap between the $i$-th and $j$-th gaussian components, and $K$is the number of mixture components.

Finally, DGMM uses epistemic uncertainty to adaptively compensate the GMM mean. A compensation branch computes an additional mean term $\mu_{an}$ from the decoding feature $H$, while $\varphi_{epi}$ controls the compensation strength through a gating weighted function. The final calibrated mean is:

$$\mu_{out} = \underbrace{\emptyset(W_{mu_a}H + b_{mu_a})}_{\mu_{an}} \odot \underbrace{\theta(W_{epi2}(W_{epi1}\varphi_{epi} + b_{epi1}) + b_{epi2})}_{Epistemic\ weight} + \mu, \tag{23}$$

where $\mu_{an}$ denotes the compensation term for high-value OD flows, $\odot$ denotes element-wise multiplication, and $\emptyset(\cdot)$ denotes the Sigmoid activation function. $W_*$ and $b_*$ $(*\in epi1, epi2)$ are learnable parameters for the compensation and gating branches. The calibrated mean $\mu_{out}$ is used as the final OD flow estimate.

**Loss functions and training process**

ODIN is trained to learn both the predictive distribution and the expected magnitude of future OD flows. The overall objective consists of three terms: negative log-likelihood (NLL) loss, L1 regularization and entropy regularization on mixture coefficients.

**Negative Log-Likelihood Loss:** Since DGMM outputs a Gaussian mixture distribution for each OD flow, the primary objective is to maximize the likelihood of the observed ground-truth flows. Equivalently, we minimize the negative log-likelihood $\mathcal{L}_{nll}$:

$$\mathcal{L}_{nll} = -\frac{1}{N}\sum_{i=1}^{N} log(\sum_{k=1}^{K}\pi_i^k \frac{1}{\sqrt{2\pi(\sigma_i^k)^2}} exp\left(-\frac{(Y_i - \mu_i^k)^2}{2(\sigma_i^k)^2}\right)), \tag{24}$$

where $Y_i$ denotes the ground-truth OD flow sample, and $\pi_i^k$, $\mu_i^k$ and $\sigma_i^k$ denote the mixture coefficient, mean and standard deviation of the $k$-th Gaussian component.

**L1 Regularization:** Likelihood optimization alone may produce unstable mean estimates under sparse and long-tailed OD distributions. We therefore add an L1 term[70] between the observed flow and the expectation of the mixture distribution.

$$\mathcal{L}_{L1} = \frac{1}{N}\sum_{i=1}^{N}\left|Y_i - (\sum_{k=1}^{K}\pi_i^k \mu_{out_i}^k)\right|, \tag{25}$$

which provides direct supervision on the predicted flow magnitude and improves robustness to extreme errors.

**Entropy Regularization on Mixture Coefficients**: To prevent the GMM from collapsing to a single dominant component, we impose entropy regularization on the mixture coefficients:

$$\mathcal{L}_{en} = \frac{1}{N}\sum_{i=1}^{N}\sum_{k=1}^{K}\pi_i^k \log(\pi_i^k), \tag{26}$$

which encourages different Gaussian components to capture different flow regimes.

The final training objective is:

$$\mathcal{L} = \mathcal{L}_{nll} + \lambda_1\mathcal{L}_{L1} + \lambda_2\mathcal{L}_{en}, \tag{27}$$

where $\lambda_1$ and $\lambda_2$ are regularization weight coefficients. To further stabilize training, we maintain an exponential moving average[71] (ema) of the model parameters. After each gradient update, the smoothed parameters $\Theta'$ are updated as:

$$\Theta' \leftarrow s\Theta' + (1-s)\Theta, \tag{28}$$

where $\Theta$ denotes the current model parameters and $s$ is the smoothing factor. The ema parameters are used to reduce abrupt parameter fluctuations during optimization. The full training procedure is summarized in Extending Data Table 2.

**Datasets**

We evaluated ODIN using twelve dynamic mobility datasets from cities in the United States and China (Extending Data Fig. 2). The U.S. datasets were constructed from public taxi and bike-sharing trip records in New York City and Chicago. For New York City, trips were aggregated by taxi zones, whereas for Chicago, trips were aggregated by community areas. For each U.S. city and mobility mode, we constructed two evaluation settings: a core-region setting containing high-volume spatial units and an ordinary-region setting containing all non-redundant spatial units. The China datasets were constructed from anonymized mobile-phone mobility records in Beijing, Shanghai, Guangzhou and Shenzhen, and were aggregated by subdistricts.

All raw mobility records were converted into temporally indexed OD matrices, where each entry denotes the number of movements from an origin spatial unit to a destination spatial unit within a given aggregation interval. Because the native temporal resolution differs across data sources, the U.S. datasets were organized at a 1-hour interval, whereas the China datasets were organized at a 2-hour interval. Accordingly, each U.S. dataset contains 8,784 temporal snapshots, and each China dataset contains 4,392 temporal snapshots, covering the full year of 2024.

To construct node-level aggregated dynamics, we derived three time-varying signals from the OD matrices: inflow, outflow and total flow. Inflow measures the number of trips arriving at each spatial unit, outflow measures the number of trips departing from each spatial unit, and total flow is the sum of the two. These aggregated signals serve as the input to the proposed coarse-to-fine prediction task, whereas the full OD matrices are used only as prediction targets.

Because dynamic OD matrices are highly sparse, spatial units with extremely limited mobility activity may introduce unstable training and evaluation signals. We therefore applied a systematic filtering procedure. Spatial units were grouped into three categories according to their average hourly mobility volume: core units, whose average hourly flow exceeds the regional mean; ordinary units, whose average hourly total flow exceeds one; and redundant units, which do not satisfy either criterion. Redundant units were removed from model training and evaluation. Based on the remaining units, we constructed two evaluation settings for each city and mobility mode: a core-region setting containing only high-volume units, and an ordinary-region setting containing all non-redundant units. The detailed dataset statistics after filtering are provided in Extended Data Table 3.

For each dataset, the time series was split chronologically into training, validation and test sets with a ratio of 7:1:2. To keep the learning setting consistent across temporal resolutions, all models used 12 historical time steps to predict OD flows for the subsequent 4 time steps. Z-score normalization was applied to the input node-level dynamics using statistics computed from the training set.

Additionally, to evaluate OD prediction performance, we employ three regression metrics and two metrics of uncertainty estimation, including RMSE, MAE, WAPE, CRPS, JSD. For CRPS computation, the number of samples was set to 50. These metrics are calculated as follows:

$$RMSE = \sqrt{\frac{1}{n}\sum_{i=1}^{n}\left(Y_i - \hat{Y}_i\right)^2}, \tag{29}$$

$$MAE = \frac{1}{n}\sum_{i=1}^{n}\left|Y_i - \hat{Y}_i\right|, \tag{30}$$

$$WAPE = \frac{\sum_{i=1}^{n}\left|Y_i - \hat{Y}_i\right|}{\sum_{i=1}^{n}|Y_i|}, \tag{31}$$

$$CRPS = \int (f(x) - \mathbf{1}(x \geq y))^2 dx \tag{32}$$

$$JSD = \sqrt{\frac{1}{2}(D_{KL}\left(P \parallel \left(\frac{P+Q}{2}\right)\right) + D_{KL}\left(\left(\frac{P+Q}{2}\right) \parallel P\right))} \tag{33}$$

where $Y_i$ denotes the ground-truth OD flow, $\hat{Y}_i$ denotes the predicted OD flow, $P$ denotes the ground-truth OD flow distribution, $Q$ denotes the predicted OD flow distribution, $f(\cdot)$ is the predicted cumulative distribution function, and $\mathbf{1}$ denotes the indicator function.

# Figures and tables

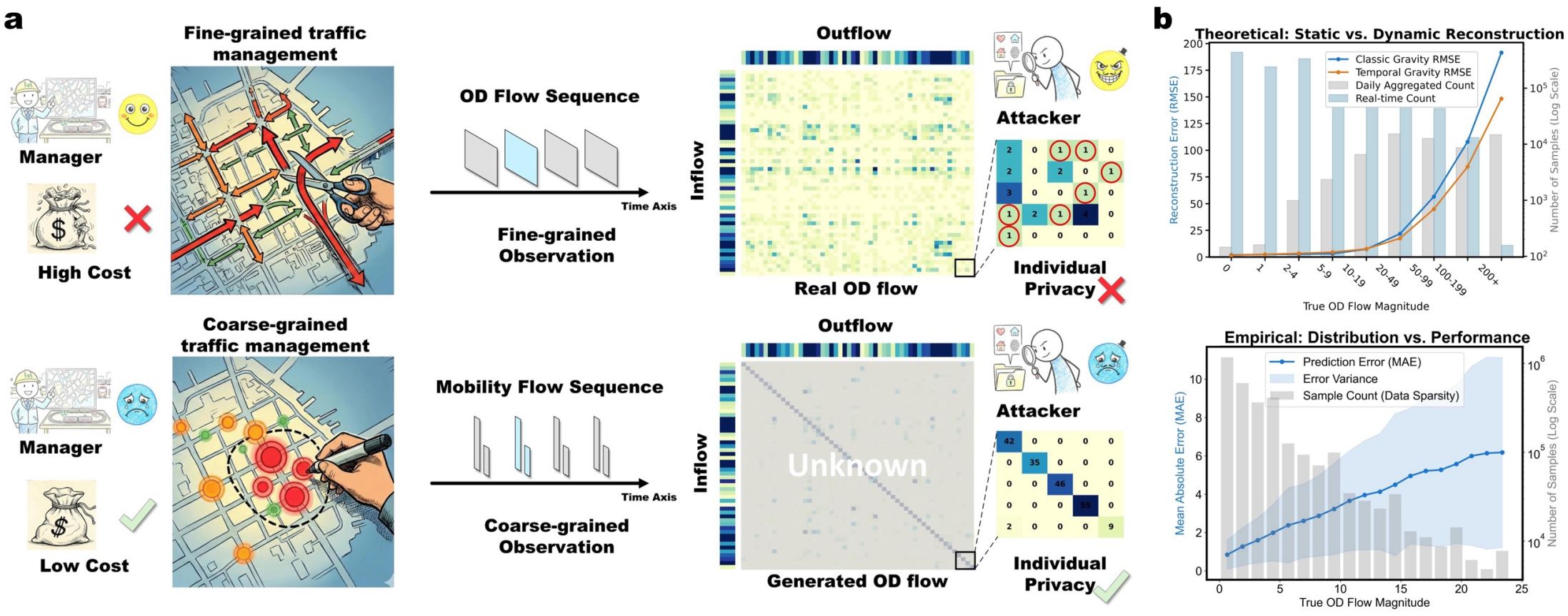

**Fig. 1.** (a) Fine-grained OD observations provide directional mobility interactions needed for traffic management, but obtaining such data often requires trajectory-level acquisition, which is costly and exposes sparse origin–destination records to potential privacy risks. In contrast, area-level aggregated inflow and outflow observations are easier to collect and less privacy-invasive, but the underlying OD matrix is unobserved. The proposed task aims to infer future fine-grained OD interactions from these spatially aggregated mobility dynamics. (b) Increasing spatially aggregated mobility flow expands the difficulty for OD dynamics prediction: 1. Real-time OD flows differ substantially from daily aggregated OD flows, showing stronger sparsity and heavier imbalance across flow magnitudes. These temporal distributional changes weaken the dense-interaction assumptions of classical gravity models, leading to amplified inference errors in dynamic OD settings. To examine this limitation, we augment the classical gravity formulation with time-dependent weights using time step index, allowing origin-destination interactions to vary across temporal states rather than being governed by a single static interaction pattern. This temporal extension reduces errors, particularly for high-flow OD regimes; 2. Empirical analysis using results of STAEformer[45] at NYC taxi dataset. As the OD flow magnitude increases, the variation of the prediction error expands. The empirical distribution of OD flows further reveals the difficulty of dynamic inference. High-magnitude interactions are rare but associated with larger prediction errors and higher uncertainty, emphasizing the need for models that explicitly account for temporal OD dynamics.

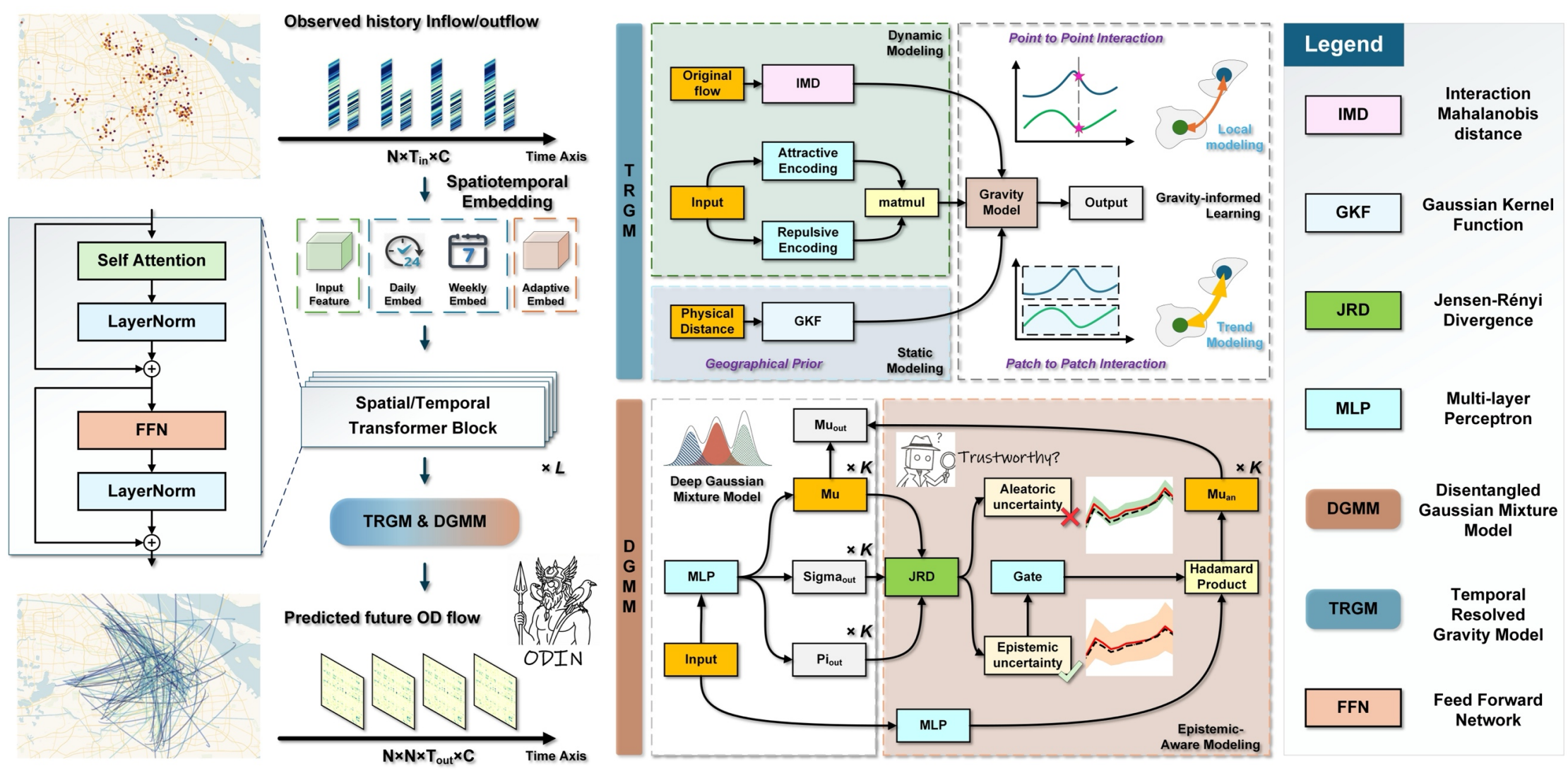

**Fig. 2.** The overall framework of proposed ODIN. The framework adopts a Transformer paradigm integrated with gravity-informed learning and uncertainty learning modules, comprising three core components: a Spatiotemporal Transformer, a Temporal Resolved Gravity Model (TRGM), and a Disentangled Gaussian Mixture Model (DGMM).

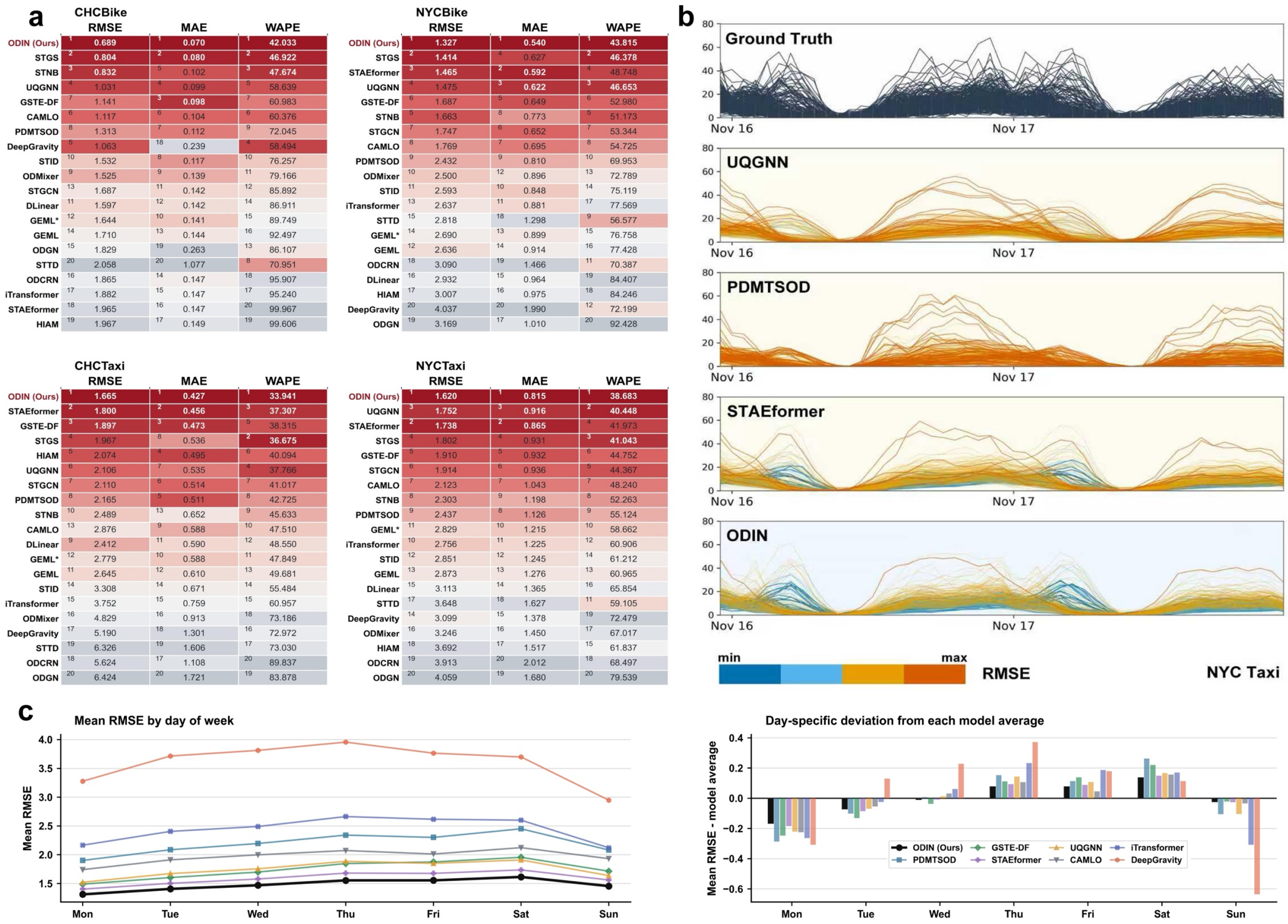

**Fig. 3**. (a) Overall prediction performance across taxi and bike-sharing datasets in New York City and Chicago. Each table reports RMSE, MAE and WAPE for ODIN and baseline models, with darker colors indicating better performance. ODIN achieves consistently strong performance across datasets. For results from other datasets, please refer to Figs. S2 and S3. (b) Temporal evolution of OD flow observations and predicted results on the NYC Taxi dataset. Each curve represents an OD pair, and color indicates RMSE magnitude. Compared with representative baselines, ODIN better preserves the temporal structure of observed mobility dynamics and produces fewer high-error trajectories. (c) Day-of-week error distribution on the NYC Taxi dataset. The left panel reports mean RMSE from Monday to Sunday, and the right panel shows each day's deviation from the model average. ODIN maintains low and stable errors across weekdays and weekends, with the smallest weekend fluctuation among compared models.

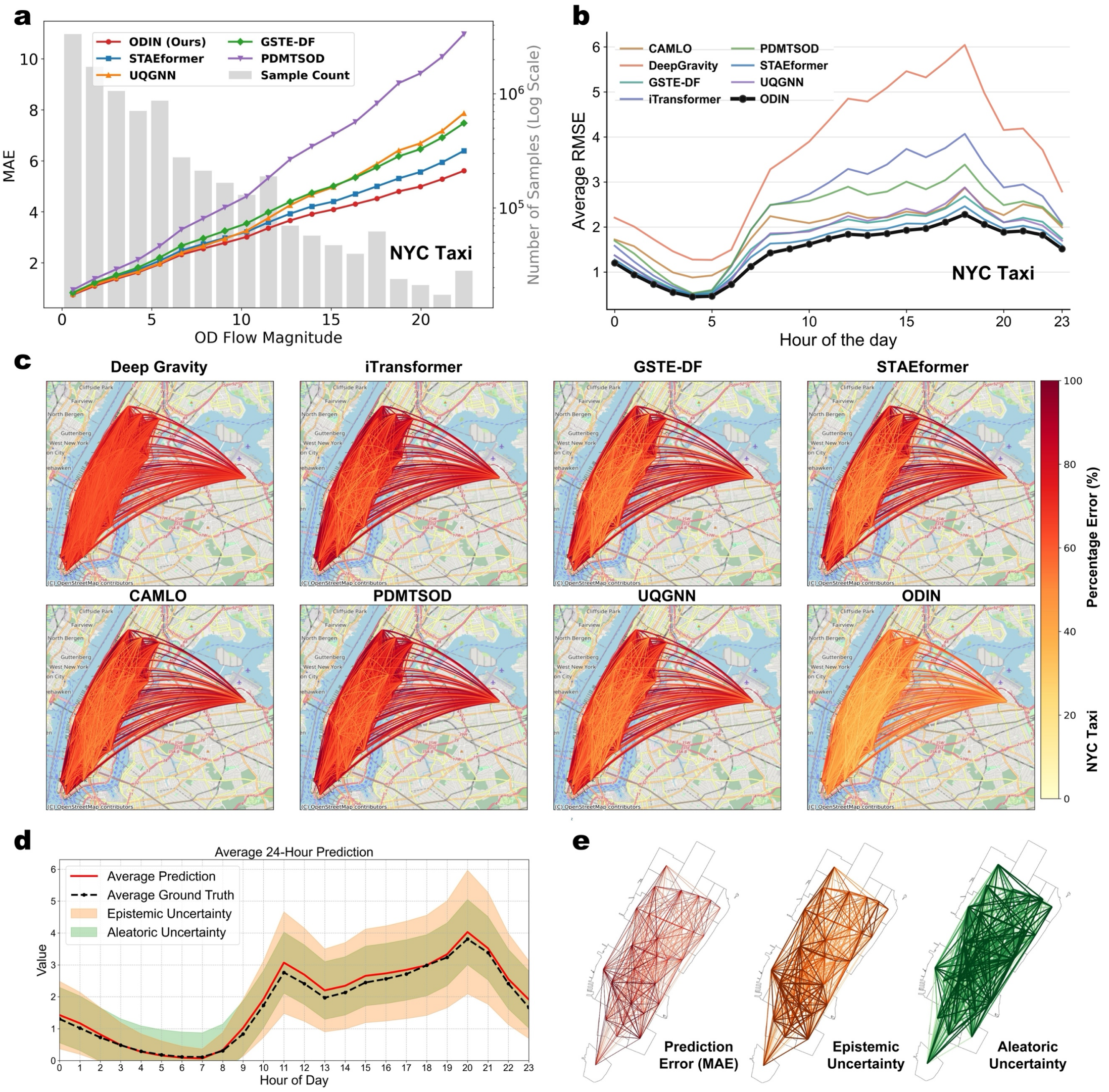

**Fig. 4**. Error and uncertainty analysis on the NYC Taxi dataset. (a) The bars show the sample count of OD flows at different magnitudes on a logarithmic scale, and the lines show MAE for different models. ODIN exhibits a slower error increase as flow magnitude grows, indicating better reconstruction of sparse high-value flows. (b) Temporal error distribution. Average hourly RMSE shows that ODIN maintains lower prediction errors across the daily mobility cycle, including both low-demand and peak-demand periods. (c) Spatial error distribution. OD links are visualized by percentage error, showing that ODIN produces fewer high-error links and a more spatially coherent error pattern than competing methods. (d) Temporal uncertainty decomposition. ODIN's predicted mean follows the average 24-hour ground truth, while epistemic and aleatoric uncertainty describe different temporal sources of prediction uncertainty. (e) Spatial uncertainty decomposition. The spatial distributions of prediction error, epistemic uncertainty and aleatoric uncertainty show that epistemic uncertainty better aligns with difficult OD links, supporting its use for reliability-aware calibration.

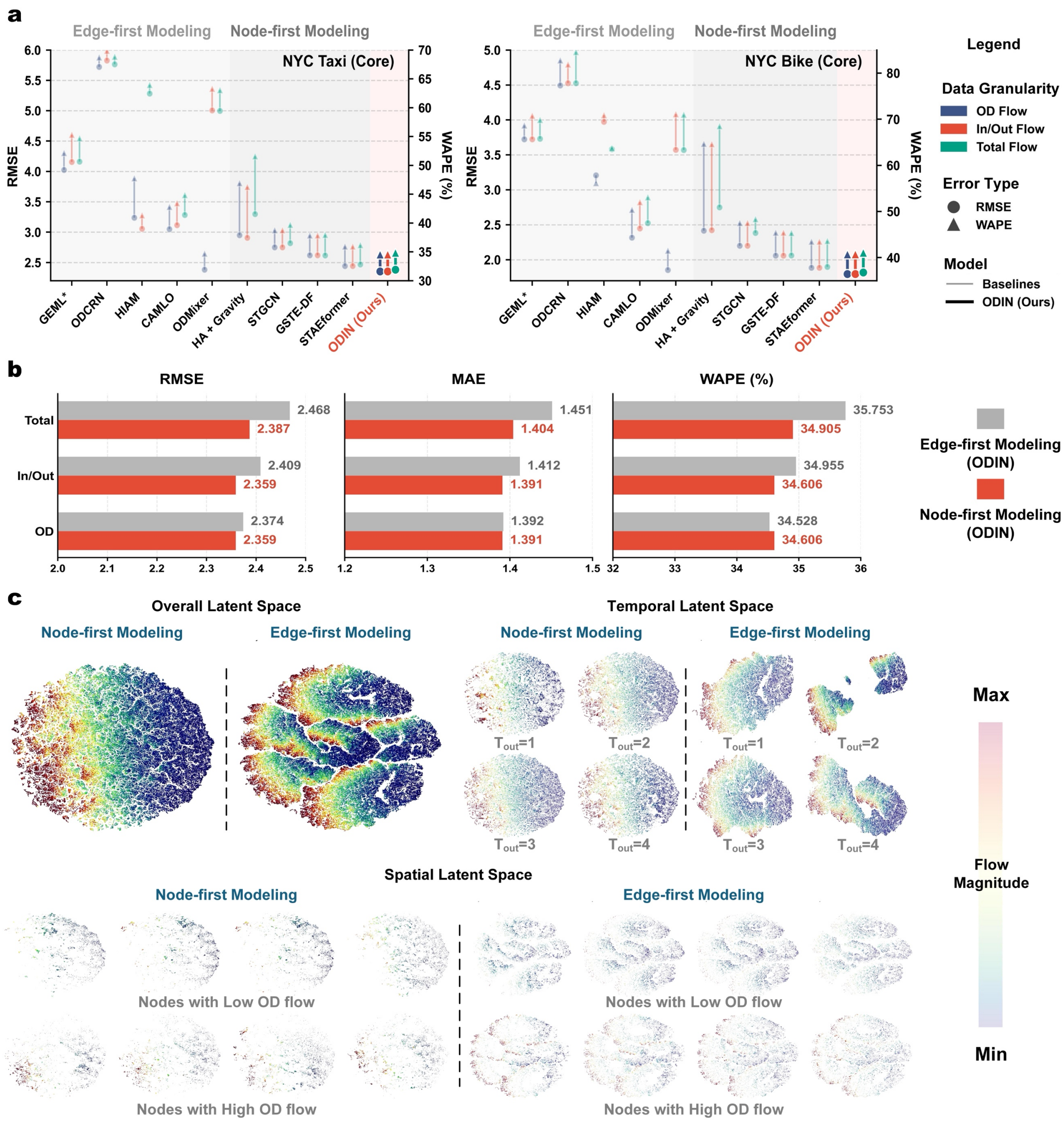

**Fig. 5.** (a) Performance of edge-first and node-first models under different input granularities on NYC Taxi Core and NYC Bike Core datasets. Circles denote RMSE and triangles denote WAPE. Node-first models show stronger and more stable inference performance, especially when only aggregated inflow/outflow or total-flow observations are available. (b) Controlled comparison between the original node-first ODIN and an edge-first variant of ODIN under OD, inflow/outflow and total-flow inputs. The node-first design consistently achieves lower RMSE, MAE and WAPE, indicating that the performance advantage is not only caused by input information but also by modeling order. (c) Latent-space visualization of node-first and edge-first representations. Node-first modeling produces a smoother and more compact overall latent space[56], clearer temporal organization across prediction horizons and stronger spatial separation between low-flow and high-flow nodes. Edge-first modeling yields more fragmented representations. These patterns suggest that learning spatial-unit mobility states before reconstructing pairwise interactions better preserves spatial heterogeneity and aligns with the generation-before-assignment logic in transportation planning. In the spatial dimension, we randomly select four nodes each from the top and bottom 10% by flow volume for visualization.

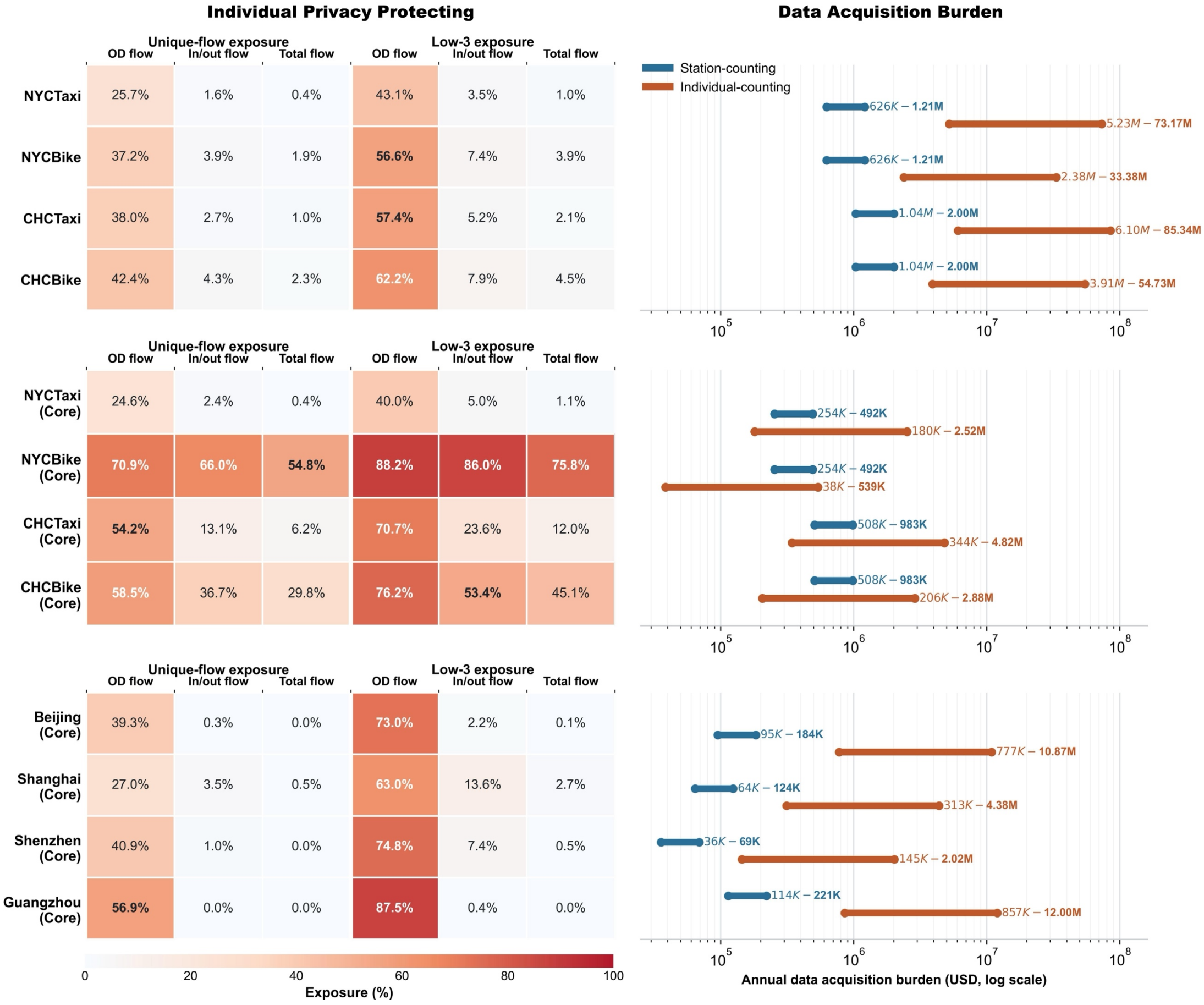

**Fig. 6.** Privacy exposure and acquisition burden of OD-level and aggregated mobility observations. Left, privacy exposure measured by cell-level uniqueness and low-3 risk across OD flow, inflow/outflow and total-flow observations. OD-level observations are evaluated over origin–destination–time cells, whereas aggregated observations are evaluated over node–time cells. Aggregated inflow/outflow and total-flow observations substantially reduce the proportion of unique and low-count mobility records, indicating lower exposure of sparse and potentially distinguishable movements. Right, annual data acquisition burden under area-level aggregated sensing and individual-level trajectory acquisition. Area-level costs are estimated from annualized traffic-count prices, whereas individual-counting costs are estimated from trajectory-record acquisition scenarios. Across taxi and bike-sharing datasets, aggregated sensing requires lower annual acquisition costs than trajectory-level acquisition, supporting the practical value of coarse-to-fine OD flow prediction.

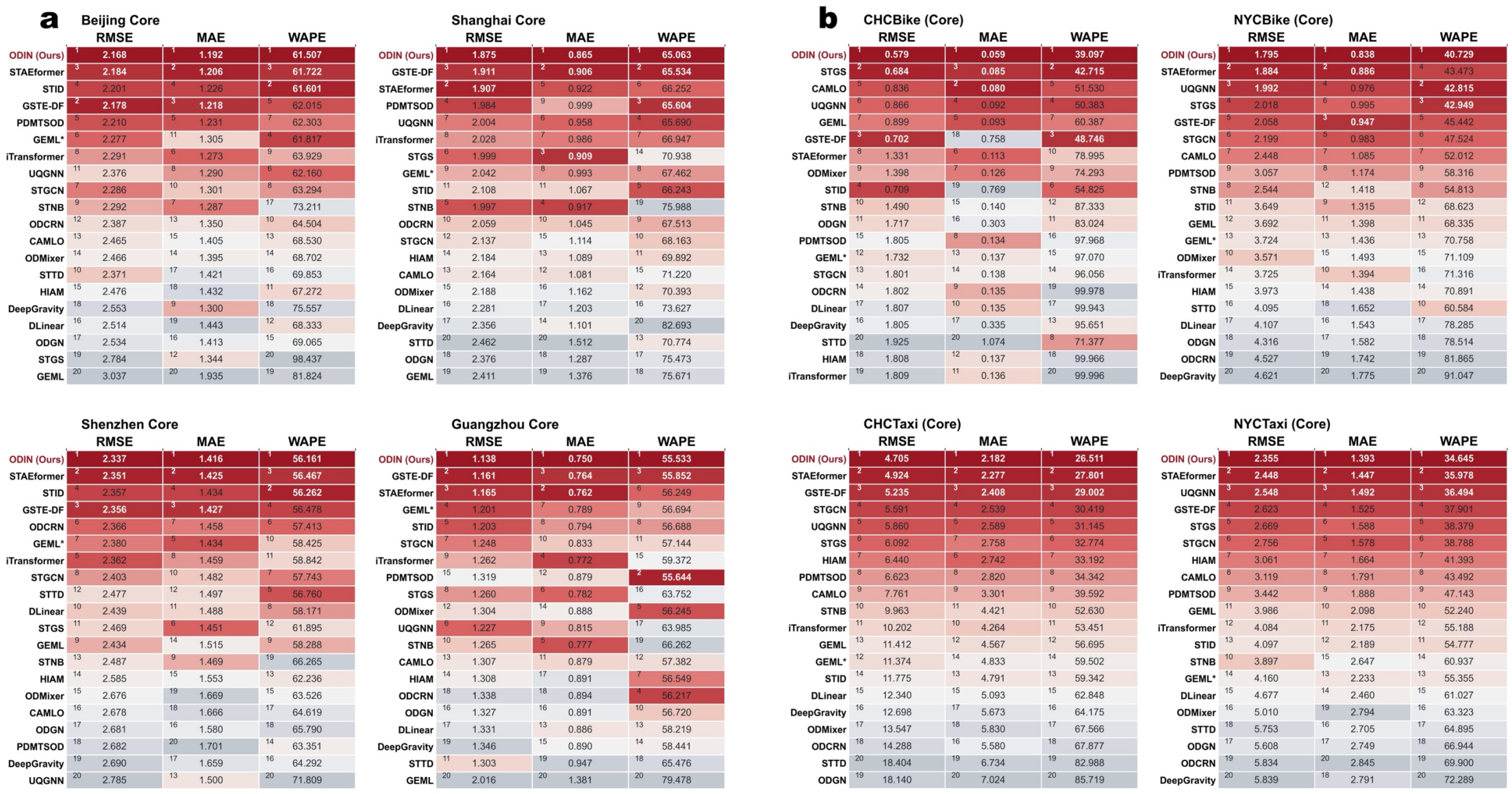

**a**

| Beijing Core | RMSE | MAE | WAPE |
|---|---|---|---|
| ODIN (Ours) | 2.168 | 1.192 | 61.507 |
| STAEformer | 2.184 | 1.206 | 61.722 |
| STID | 2.201 | 1.226 | 61.601 |
| GSTE-DF | 2.178 | 1.218 | 62.015 |
| PDMTSOD | 2.210 | 1.231 | 62.303 |
| GEML* | 2.277 | 1.305 | 61.817 |
| iTransformer | 2.291 | 1.273 | 63.929 |
| UQGNN | 2.376 | 1.290 | 62.160 |
| STGCN | 2.286 | 1.301 | 63.294 |
| STNB | 2.292 | 1.287 | 73.211 |
| ODCRN | 2.387 | 1.350 | 64.504 |
| CAMLO | 2.465 | 1.405 | 68.530 |
| ODMixer | 2.466 | 1.395 | 68.702 |
| STTD | 2.371 | 1.421 | 69.853 |
| HIAM | 2.476 | 1.432 | 67.272 |
| DeepGravity | 2.553 | 1.300 | 75.557 |
| DLinear | 2.514 | 1.443 | 68.333 |
| ODGN | 2.534 | 1.413 | 69.065 |
| STGS | 2.784 | 1.344 | 98.437 |
| GEML | 3.037 | 1.935 | 81.824 |

| Shanghai Core | RMSE | MAE | WAPE |
|---|---|---|---|
| ODIN (Ours) | 1.875 | 0.865 | 65.063 |
| GSTE-DF | 1.911 | 0.906 | 65.534 |
| STAEformer | 1.907 | 0.922 | 66.252 |
| PDMTSOD | 1.984 | 0.999 | 65.604 |
| UQGNN | 2.004 | 0.958 | 65.690 |
| iTransformer | 2.028 | 0.986 | 66.947 |
| STGS | 1.999 | 0.909 | 70.938 |
| GEML* | 2.042 | 0.993 | 67.462 |
| STID | 2.108 | 1.067 | 66.243 |
| STNB | 1.997 | 0.917 | 75.988 |
| ODCRN | 2.059 | 1.045 | 67.513 |
| STGCN | 2.137 | 1.114 | 68.163 |
| HIAM | 2.184 | 1.089 | 69.892 |
| CAMLO | 2.164 | 1.081 | 71.220 |
| ODMixer | 2.188 | 1.162 | 70.393 |
| DLinear | 2.281 | 1.203 | 73.627 |
| DeepGravity | 2.356 | 1.101 | 82.693 |
| STTD | 2.462 | 1.512 | 70.774 |
| ODGN | 2.376 | 1.287 | 75.473 |
| GEML | 2.411 | 1.376 | 75.671 |

**b**

| CHCBike (Core) | RMSE | MAE | WAPE |
|---|---|---|---|
| ODIN (Ours) | 0.579 | 0.059 | 39.097 |
| STGS | 0.684 | 0.085 | 42.715 |
| CAMLO | 0.836 | 0.080 | 51.530 |
| UQGNN | 0.866 | 0.092 | 50.383 |
| GEML | 0.899 | 0.093 | 60.387 |
| GSTE-DF | 0.702 | 0.758 | 48.746 |
| STAEformer | 1.331 | 0.113 | 78.995 |
| ODMixer | 1.398 | 0.126 | 74.293 |
| STID | 0.709 | 0.769 | 54.825 |
| STNB | 1.490 | 0.140 | 87.333 |
| ODGN | 1.717 | 0.303 | 83.024 |
| PDMTSOD | 1.805 | 0.134 | 97.968 |
| GEML* | 1.732 | 0.137 | 97.070 |
| STGCN | 1.801 | 0.138 | 96.056 |
| ODCRN | 1.802 | 0.135 | 99.978 |
| DLinear | 1.807 | 0.135 | 99.943 |
| DeepGravity | 1.805 | 0.335 | 95.651 |
| STTD | 1.925 | 1.074 | 71.377 |
| HIAM | 1.808 | 0.137 | 99.966 |
| iTransformer | 1.809 | 0.136 | 99.996 |

| NYCBike (Core) | RMSE | MAE | WAPE |
|---|---|---|---|
| ODIN (Ours) | 1.795 | 0.838 | 40.729 |
| STAEformer | 1.884 | 0.886 | 43.473 |
| UQGNN | 1.992 | 0.976 | 42.815 |
| STGS | 2.018 | 0.995 | 42.949 |
| GSTE-DF | 2.058 | 0.947 | 45.442 |
| STGCN | 2.199 | 0.983 | 47.524 |
| CAMLO | 2.448 | 1.085 | 52.012 |
| PDMTSOD | 3.057 | 1.174 | 58.316 |
| STNB | 2.544 | 1.418 | 54.813 |
| STID | 3.649 | 1.315 | 68.623 |
| GEML | 3.692 | 1.398 | 68.335 |
| GEML* | 3.724 | 1.436 | 70.758 |
| ODMixer | 3.571 | 1.493 | 71.109 |
| iTransformer | 3.725 | 1.394 | 71.316 |
| HIAM | 3.973 | 1.438 | 70.891 |
| STTD | 4.095 | 1.652 | 60.584 |
| DLinear | 4.107 | 1.543 | 78.285 |
| ODGN | 4.316 | 1.582 | 78.514 |
| ODCRN | 4.527 | 1.742 | 81.865 |
| DeepGravity | 4.621 | 1.775 | 91.047 |

| Shenzhen Core | RMSE | MAE | WAPE |
|---|---|---|---|
| ODIN (Ours) | 2.337 | 1.416 | 56.161 |
| STAEformer | 2.351 | 1.425 | 56.467 |
| STID | 2.357 | 1.434 | 56.262 |
| GSTE-DF | 2.356 | 1.427 | 56.478 |
| ODCRN | 2.366 | 1.458 | 57.413 |
| GEML* | 2.380 | 1.434 | 58.425 |
| iTransformer | 2.362 | 1.459 | 58.842 |
| STGCN | 2.403 | 1.482 | 57.743 |
| STTD | 2.477 | 1.497 | 56.760 |
| DLinear | 2.439 | 1.488 | 58.171 |
| STGS | 2.469 | 1.451 | 61.895 |
| GEML | 2.434 | 1.515 | 58.288 |
| STNB | 2.487 | 1.469 | 66.265 |
| HIAM | 2.585 | 1.553 | 62.236 |
| ODMixer | 2.676 | 1.669 | 63.526 |
| CAMLO | 2.678 | 1.666 | 64.619 |
| ODGN | 2.681 | 1.580 | 65.790 |
| PDMTSOD | 2.682 | 1.701 | 63.351 |
| DeepGravity | 2.690 | 1.659 | 64.292 |
| UQGNN | 2.785 | 1.500 | 71.809 |

| Guangzhou Core | RMSE | MAE | WAPE |
|---|---|---|---|
| ODIN (Ours) | 1.138 | 0.750 | 55.533 |
| GSTE-DF | 1.161 | 0.764 | 55.852 |
| STAEformer | 1.165 | 0.762 | 56.249 |
| GEML* | 1.201 | 0.789 | 56.694 |
| STID | 1.203 | 0.794 | 56.688 |
| STGCN | 1.248 | 0.833 | 57.144 |
| iTransformer | 1.262 | 0.772 | 59.372 |
| PDMTSOD | 1.319 | 0.879 | 55.644 |
| STGS | 1.260 | 0.782 | 63.752 |
| ODMixer | 1.304 | 0.888 | 56.245 |
| UQGNN | 1.227 | 0.815 | 63.985 |
| STNB | 1.265 | 0.777 | 66.262 |
| CAMLO | 1.307 | 0.879 | 57.382 |
| HIAM | 1.308 | 0.891 | 56.549 |
| ODCRN | 1.338 | 0.894 | 56.217 |
| ODGN | 1.327 | 0.891 | 56.720 |
| DLinear | 1.331 | 0.886 | 58.219 |
| DeepGravity | 1.346 | 0.890 | 58.441 |
| STTD | 1.303 | 0.947 | 65.476 |
| GEML | 2.016 | 1.381 | 79.478 |

| CHCTaxi (Core) | RMSE | MAE | WAPE |
|---|---|---|---|
| ODIN (Ours) | 4.705 | 2.182 | 26.511 |
| STAEformer | 4.924 | 2.277 | 27.801 |
| GSTE-DF | 5.235 | 2.408 | 29.002 |
| STGCN | 5.591 | 2.539 | 30.419 |
| UQGNN | 5.860 | 2.589 | 31.145 |
| STGS | 6.092 | 2.758 | 32.774 |
| HIAM | 6.440 | 2.742 | 33.192 |
| PDMTSOD | 6.623 | 2.820 | 34.342 |
| CAMLO | 7.761 | 3.301 | 39.592 |
| STNB | 9.963 | 4.421 | 52.630 |
| iTransformer | 10.202 | 4.264 | 53.451 |
| GEML | 11.412 | 4.567 | 56.695 |
| GEML* | 11.374 | 4.833 | 59.502 |
| STID | 11.775 | 4.791 | 59.342 |
| DLinear | 12.340 | 5.093 | 62.848 |
| DeepGravity | 12.698 | 5.673 | 64.175 |
| ODMixer | 13.547 | 5.830 | 67.566 |
| ODCRN | 14.288 | 5.580 | 67.877 |
| STTD | 18.404 | 6.734 | 82.988 |
| ODGN | 18.140 | 7.024 | 85.719 |

| NYCTaxi (Core) | RMSE | MAE | WAPE |
|---|---|---|---|
| ODIN (Ours) | 2.355 | 1.393 | 34.645 |
| STAEformer | 2.448 | 1.447 | 35.978 |
| UQGNN | 2.548 | 1.492 | 36.494 |
| GSTE-DF | 2.623 | 1.525 | 37.901 |
| STGS | 2.669 | 1.588 | 38.379 |
| STGCN | 2.756 | 1.578 | 38.788 |
| HIAM | 3.061 | 1.664 | 41.393 |
| CAMLO | 3.119 | 1.791 | 43.492 |
| PDMTSOD | 3.442 | 1.888 | 47.143 |
| GEML | 3.986 | 2.098 | 52.240 |
| iTransformer | 4.084 | 2.175 | 55.188 |
| STID | 4.097 | 2.189 | 54.777 |
| STNB | 3.897 | 2.647 | 60.937 |
| GEML* | 4.160 | 2.233 | 55.355 |
| DLinear | 4.677 | 2.460 | 61.027 |
| ODMixer | 5.010 | 2.794 | 63.323 |
| STTD | 5.753 | 2.705 | 64.895 |
| ODGN | 5.608 | 2.749 | 66.944 |
| ODCRN | 5.834 | 2.845 | 69.900 |
| DeepGravity | 5.839 | 2.791 | 72.289 |

**Extending Data Fig. 1.** Overall performance across additional urban mobility datasets. (a) Prediction performance on four core-area mobility datasets from Beijing, Shanghai, Shenzhen and Guangzhou. (b) Prediction performance on taxi and bike-sharing core-region datasets from Chicago and New York City. Each table reports RMSE, MAE and WAPE for ODIN and baseline models. Across different cities, transport modes and spatial settings, ODIN consistently achieves top or near-top performance while using only spatially aggregated mobility dynamics as input.

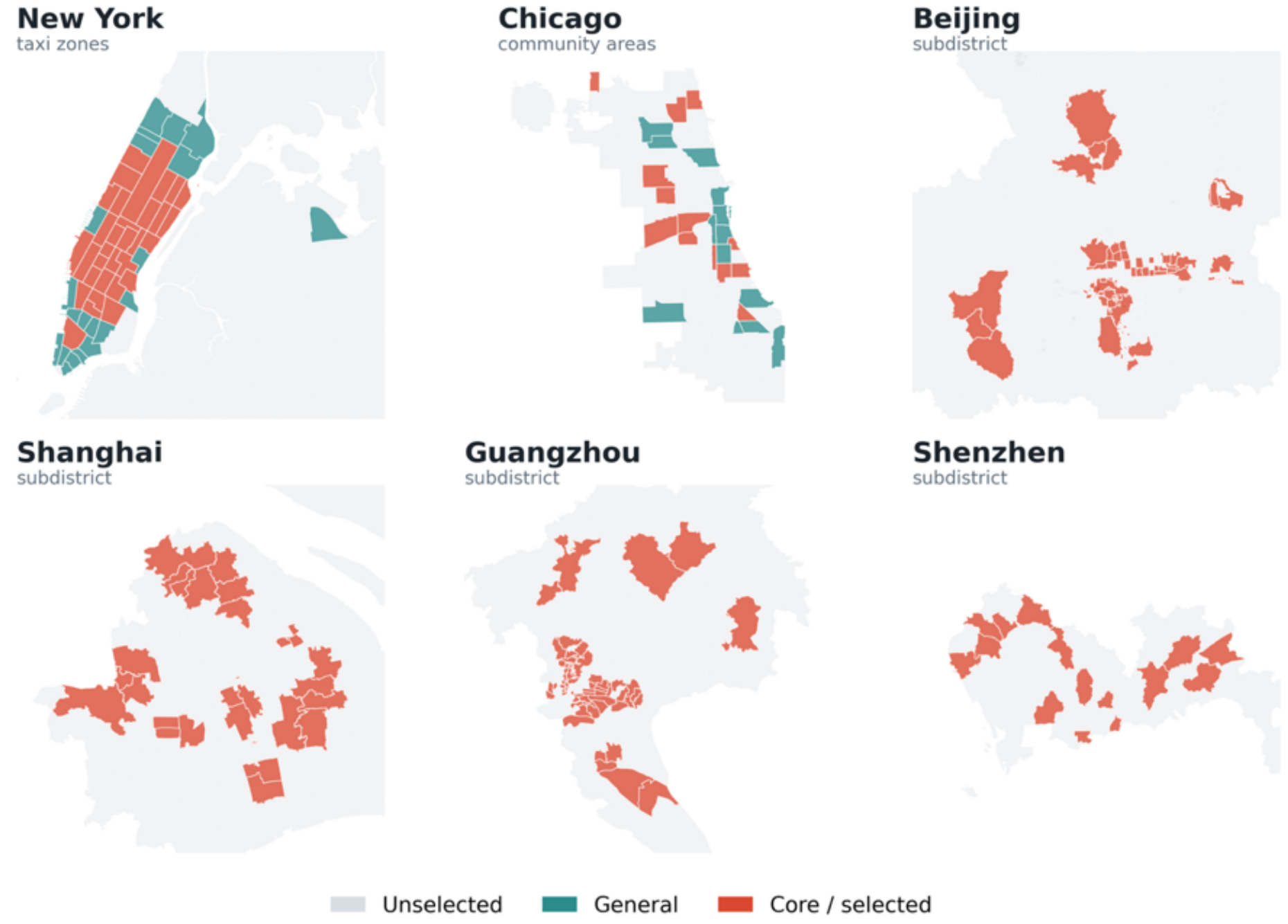

**Extending Data Fig. 2.** Spatial coverage of the mobility datasets in New York, Chicago, Beijing, Shanghai, Guangzhou and Shenzhen. New York uses taxi zones, Chicago uses community areas, and Chinese cities use subdistricts as spatial units. Grey denotes unselected areas, teal denotes general study regions and orange denotes core or selected regions for evaluation.

**Extending Data Table 1** Trustworthy performance of different uncertainty models. ** mean that the values are too large to be negligible.

| Datasets \ Benchmarks | NYCBike (Core) | | NYCTaxi (Core) | | NYCBike | | NYCTaxi | |
|---|---|---|---|---|---|---|---|---|
| | CRPS | JSD | CRPS | JSD | CRPS | JSD | CRPS | JSD |
| STNB | ** | 0.484 | ** | 0.409 | ** | 0.422 | ** | 0.465 |
| STTD | ** | 0.562 | ** | 0.456 | ** | 0.620 | ** | 0.550 |
| STGS | 0.876 | 0.299 | 1.381 | 0.230 | 0.541 | 0.317 | 0.832 | 0.279 |
| UQGNN | 1.931 | 0.586 | 3.197 | 0.486 | 1.257 | 0.659 | 2.018 | 0.605 |
| **ODIN(Ours)** | **0.692** | **0.273** | **1.111** | **0.209** | **0.458** | **0.303** | **0.679** | **0.259** |
| **Datasets \ Benchmarks** | **NYCBike (Core)** | | **NYCTaxi (Core)** | | **NYCBike** | | **NYCTaxi** | |
| | **CRPS** | **JSD** | **CRPS** | **JSD** | **CRPS** | **JSD** | **CRPS** | **JSD** |
| STNB | ** | 0.595 | ** | 0.420 | ** | 0.427 | ** | 0.393 |
| STTD | ** | 0.800 | ** | 0.525 | ** | 0.796 | ** | 0.695 |
| STGS | 0.078 | 0.334 | 2.499 | 0.192 | 0.073 | 0.311 | 0.498 | 0.281 |
| UQGNN | 0.265 | 0.818 | 9.322 | 0.580 | 0.288 | 0.814 | 1.856 | 0.775 |
| **ODIN(Ours)** | **0.050** | **0.270** | **1.706** | **0.160** | **0.058** | **0.287** | **0.362** | **0.251** |

**Extending Data Table 2** The pseudocode of training ODIN.

**Algorithm** The training process of ODIN.

1: **Input:**
2: Node-level inflow data and outflow data: $X_{in} = \{X_{t-q+1}^{in}, X_{t-q+2}^{in}, \ldots, X_t^{in}\}$, $X_{out} = \{X_{t-q+1}^{out}, X_{t-q+2}^{out}, \ldots, X_t^{out}\}$,
3: Time-stamps data of time-of-day and day-of-week: $S_H^s = \{S_{t-q+1}^s, S_{t-q+2}^s, \ldots, S_t^s\}$, $S_H^w = \{S_{t-q+1}^w, S_{t-q+2}^w, \ldots, S_t^w\}$
4: The initialized node-level dynamics embedding: $\mathcal{F}_{nde}(\cdot)$
5: The initialized spatial transformer module: $\mathcal{F}_S(\cdot)$
6: The initialized temporal transformer module: $\mathcal{F}_T(\cdot)$
7: The initialized temporal resolved gravity model: $\mathcal{F}_{TRGM}(\cdot)$
8: The initialized disentangled gaussian mixture model: $\mathcal{F}_{DGMM}(\cdot)$
9: Hyperparameter: Predicted snapshots: $p$, Historical snapshots: $q$, Learning rate $r$, Smoothing factor $s$
10: Ground truth of OD flow: $Y_{GT} = \{Y_{t+1}, Y_{t+2}, \ldots, Y_{t+p}\}$
11: **Output:**
12: Learnable parameters of ODIN: $\Theta$
13: Predicted future OD flow: $Y_P = \{\hat{Y}_{t+1}, \hat{Y}_{t+2}, \ldots, \hat{Y}_{t+p}\}$
14: **procedure** ODIN Training
15: represent the spatiotemporal embedding feature $Z^0 \to \mathcal{F}_{ste}(X_H, S_H^s, S_H^w)$
16: **for** $l$ in $0{:}L$ **do**
17: compute the dynamic temporal correlation feature $Z_T^l \to \mathcal{F}_T^l(Z_T^{l-1})$
18: **for** $l$ in $0{:}L$ **do**
19: compute the dynamic spatial correlation feature $Z_S^l \to \mathcal{F}_S^l(Z_S^{l-1})$
20: reconstruct edge-level feature using the node-level feature $I \to \mathcal{F}_{TRGM}(Z_{ST}^L)$
21: decode the OD flow from latent feature via uncertainty modeling $Y_P \to \mathcal{F}_{DGMM}(I)$
22: calculate the loss from prediction and ground truth $Loss \to \mathcal{L}(Y_{GT}, Y_P)$
23: weights the learnable parameters $s\Theta + (1-s)\Theta'$ using exponential moving average
24: updates the learnable parameters $\Theta'$ based on their gradients and learning rate $r$
25: **end for**
26: **repeat**

27: optimize the model parameters by Adam optimizer to minimize $Loss$

28: **until** convergence

29: **return** learned parameters $\Theta$

**Extending Data Table 3** Description of human mobility datasets.

| Category | U.S. datasets | | | | China datasets | | | |
|---|---|---|---|---|---|---|---|---|
| | **New York (Core)** | **New York** | **Chicago (Core)** | **Chicago** | **Beijing (Core)** | **Shanghai (Core)** | **Guangzhou (Core)** | **Shenzhen (Core)** |
| **Time Span** | 2024.1.1-2024.12.31 | | | | | | | |
| **Time Interval** | 1 h / Snapshot | | | | 2 h / Snapshot | | | |
| **Spatial Units** | 32 | 53 | 13 | 26 | 40 | 25 | 48 | 15 |
| **Attributes** | Taxi, Bike | | | | Cell Phone | | | |
| **Zero Percentage** | Taxi: 33.465% Bike: 42.086% | Taxi: 38.944% Bike: 38.684% | Taxi: 32.667% Bike: 49.400% | Taxi: 44.838% Bike: 49.242% | 64.90% | 76.83% | 57.79% | 50.12% |
| **Source Website** | https://on.nyc.gov/Taxis; https://www.nyc.gov/html/dot/html.shtml | | https://data.cityofchicago.org/Transportation/Taxi-Trips-2024-/ajtu-isnz; https://divvybikes.com/system-data | | None | | | |
| **Spatial Grain** | Taxi Zone | | Community Area | | Subdistrict | | | |